\documentclass[twoside,11pt]{article}

\usepackage{jmlr2e}
\usepackage{graphicx}
\usepackage{pbox}
\usepackage{graphicx}
\usepackage{booktabs}
\usepackage{url}
\usepackage{color}
\usepackage{csvsimple}
\usepackage{booktabs}
\usepackage{soul,color}

\usepackage{amsmath}
\usepackage{graphicx}
\usepackage{amssymb}
\usepackage{bbm}
\usepackage{enumerate}

\usepackage{rotating}
\usepackage{graphicx}
\usepackage{caption}
\usepackage{subcaption}
\usepackage{multirow}
\newcommand{\be}{\begin{equation}}
\newcommand{\ee}{\end{equation}}
\newcommand{\ben}{\begin{equation*}}
\newcommand{\een}{\end{equation*}}
\newcommand{\ba}{\begin{array}}
\newcommand{\ea}{\end{array}}

\newcounter{Mtheorem}
{\hspace*{\fill}$\blacksquare$ \newline\end{itshape} \par}
\newcounter{trans}
{\hspace*{\fill}$\blacksquare$\newline \end{itshape} \par}

\newcounter{Mcorollary}
{\hspace*{\fill}$\blacksquare$ \newline\end{itshape} \par}
\newcounter{Mlemma}
{\hspace*{\fill}$\blacksquare$\newline\end{itshape}\par}
\newcounter{MProof}
{\hspace*{\fill}$\blacksquare$\newline\end{itshape}\par}
\newcounter{Mexample}
{\hspace*{\fill} $\blacksquare$\newline  \end{itshape}\par}
\newcounter{Mdefinition}
{\hspace*{\fill}$\blacksquare$\newline \end{itshape}\par}
\newcounter{Mproposition}
{\hspace*{\fill}$\blacksquare$\newline \end{itshape}\par}

\ShortHeadings{Should we  use the post-hoc tests based on mean-ranks?}{Benavoli and Corani and Mangili}
\firstpageno{1}

\begin{document}

\title{Should we really use post-hoc tests based on mean-ranks?}

\author{\name Alessio Benavoli\email alessio@idsia.ch \\
\name Giorgio Corani\email giorgio@idsia.ch \\
\name Francesca Mangili\email francesca@idsia.ch \\
       \addr Istituto Dalle Molle di Studi sull'Intelligenza Artificiale (IDSIA)\\
       Scuola Universitaria Professionale della Svizzera italiana (SUPSI)\\ Universit\`{a} della Svizzera italiana (USI) \\
       Manno, Switzerland\\
}

 \editor{}

\maketitle

\begin{abstract}
The statistical comparison of multiple algorithms over multiple data sets is fundamental in machine learning.
This is typically carried out by the Friedman test.
When the Friedman test rejects the null hypothesis, multiple comparisons are carried out to establish which are the significant differences among algorithms. The multiple comparisons are usually performed using the mean-ranks test. 
The aim of this technical note is to discuss the inconsistencies of the mean-ranks post-hoc test with the goal of discouraging its use in machine learning  as well as in medicine, psychology, etc..
We show that the outcome of the mean-ranks test depends on the pool of algorithms originally included in the experiment.
In other words, the outcome of the comparison between algorithms $A$ and $B$ depends also on the performance of the other algorithms included in the original experiment.
This can lead to paradoxical situations. For instance the difference between $A$ and $B$ could be declared  significant if the pool comprises algorithms $C,D,E$  and 
not significant if the pool comprises algorithms $F,G,H$.
To overcome these issues, we suggest instead to perform the multiple comparison using a test whose outcome only depends  on the two algorithms being compared, such as the sign-test or the Wilcoxon signed-rank test.
\end{abstract}

\begin{keywords}
 statistical comparison, Friedman test, post-hoc test
\end{keywords}
\section{Introduction}
The statistical comparison of multiple algorithms over multiple data sets is fundamental in machine learning;
it is typically carried out by means of a statistical test.
The recommended approach is the   Friedman test \citep{demvsar2006statistical}. Being non-parametric, it does \textit{not} require commensurability of the measures across different data sets, it does \textit{not} assume normality of the sample means and it is \textit{robust} to outliers.

When the Friedman test rejects the null hypothesis of no difference among the algorithms, 
post-hoc analysis is carried out to assess which differences are significant.
A series of pairwise comparison is performed 
adjusting the significance level 
via Bonferroni correction or other more powerful approaches \citep{demvsar2006statistical,garcia2008extension} to control the family-wise Type I error.

The mean-ranks post-hoc test \citep{McDonald1967,nemeneyi1963},
 is recommended as pairwise test for multiple comparisons in most books of nonparametric statistics: see for instance \cite[Sec. 12.2.1]{gibbons2011nonparametric}, \cite[Sec. 8.2]{kvam2007nonparametric} and \cite[Sec. 25.2]{sheskin2003handbook}. It is also commonly used in machine learning \citep{demvsar2006statistical,garcia2008extension}. 
The mean-ranks  test is based on the statistic:
$$
 z=|\bar{R}_A-\bar{R}_B|/\sqrt{\frac{m(m+1)}{6n}},
$$
where $\bar{R}_A,\bar{R}_B$ are the mean ranks (as computed by the Friedman test) of algorithms A and B, $m$ is the number of algorithms to be compared and $n$ the number of datasets.
The  mean-ranks $\bar{R}_A,\bar{R}_B$  are computed considering the performance of all the $m$ algorithms.
Thus the outcome of the comparison between $A$ and $B$ depends also on the performance of the other (m-2) algorithms included in the original experiment.
This can lead to paradoxical situations.
For instance the difference between $A$ and $B$ could be declared  \textit{significant} if the pool comprises algorithms $C,D,E$  and 
\textit{not significant} if the pool comprises algorithms $F,G,H$.
The performance of the remaining algorithms should instead be irrelevant when comparing algorithms $A$ and $B$.
This problem has been pointed out several times in the past \citep{miller1966simultaneous,gabriel1969simultaneous,Fligner1984} and also in \cite[Sec. 7.3]{hollander2013nonparametric}.
Yet it is ignored by most literature on nonparametric statistics.
However this issue should not be ignored, as it can increase the type I error when comparing two equivalent  algorithms 
and conversely decrease the power when comparing algorithms whose performance is truly different. 
In this technical note, all these inconsistencies of the mean-ranks test will be discussed in details and illustrated by means of highlighting examples  with the goal of discouraging its use in machine learning as well as in medicine, psychology, etc..


To avoid theses issues, we instead recommend to perform the pairwise comparisons of the post-hoc analysis
using  the  \textit{Wilcoxon signed-rank test} or the \textit{sign test}.
The decisions of such tests do not depend on the pool of algorithms included in the initial experiment.
It is understood that, regardless the specific test adopted for the pairwise comparisons, it is necessary to control the family-wise type I error.
This can be obtained through Bonferroni correction or through more powerful approaches \citep{demvsar2006statistical,garcia2008extension}.

Even better would be the adoption of the Bayesian methods for hypothesis testing. 
They overcome the many drawbacks \citep{demvsar2008appropriateness,goodman1999toward,kruschke2010bayesian} of the null-hypothesis significance tests.
For instance, Bayesian counterparts of the Wilcoxon and of the sign test have been presented in \citep{benavoli2014a,IDP}; 
a Bayesian approach for comparing cross-validated algorithms on multiple data sets is discussed by  \citep{ML}.

\section{Friedman test}
The performance of  multiple algorithms tested on multiple datasets can be organized in a matrix:
\begin{equation}
\label{eq:matacc}
\begin{array}{ccccc}
\multicolumn{5}{c}{~~Datasets~~}\\
\multirow{4}{*}{\begin{sideways}\textit{Algorithms}\end{sideways}} &X_{11} & X_{12} & \dots & X_{1n}\\
& X_{21} & X_{22} & \dots & X_{2n}\\
& \vdots  & \vdots & \vdots & \vdots\\
& X_{m1} & X_{m2} & \dots & X_{mn}\\
\end{array}
\end{equation} 
where $X_{ij}$ denotes the performance of the $i$-th algorithm on the $j$-th dataset (for $i=1,\dots,m$ and $j=1,\dots,n$).
The observations (performances) in different columns are assumed to be independent.
The algorithms are ranked column-by-column and each entry $X_{ij}$ is replaced by 
its rank  relative to the other observations in the $j$-th column:
\begin{equation}
\label{eq:friedmanmatrix}
\mathbf{R}=\left[  \begin{array}{cccc}
R_{11} & R_{12} & \dots & R_{1n}\\
 R_{21} & R_{22} & \dots & R_{2n}\\
 \vdots  & \vdots & \vdots & \vdots\\
 R_{m1} & R_{m2} & \dots & R_{mn}\\
\end{array}\right],
\end{equation} 
where $R_{ij}$ is the rank of the algorithm $i$ in the $j$-th dataset.
The sum of the $i$-th  row
$R_{i}= \sum_{j=1}^n R_{ij}$, $\forall ~i=1,\dots,m$, depends on how the $i$-th algorithm performs w.r.t. the other $(m-1)$ algorithms.
Under the null hypothesis of the Friedman test (no difference between the algorithms) the average value of $R_{i}$ is $n(m+1)/2$.
The statistic of the Friedman test is 
\begin{equation}
\label{eq:friedmanstat}
S=\frac{12 }{ nm(m+1)} \sum\limits_{j=1}^n \left[R_j-\frac{n(m+1)}{2}\right]^2,
\end{equation}
which under the null hypothesis has a chi-squared distribution with $m-1$ degrees of freedom. 
For $m = 2$, the Friedman test corresponds to the sign test. 

\section{Mean ranks post-hoc test}
\label{sec:direct}
If the Friedman test rejects the null hypothesis one has to establish 
which are the significant differences among the algorithms.
If all classifiers are compared to
each other, one has to perform $m(m-1)/2$ pairwise comparisons.

When performing multiple comparisons, one has to control the family-wise error rate,
namely the probability of at least one erroneous 
rejection of the null hypothesis among the $m(m-1)/2$ pairwise comparisons.
In the following example we control the family-wise error (FWER) rate through the Bonferroni correction, even though
more powerful techniques are also available \citep{demvsar2006statistical,garcia2008extension}.
However our discussion of the shortcomings of the mean-ranks test 
is valid regardless  the specific approach adopted to control the FWER. 

%
The mean-rank test claims that the $i$-th and the  $j$-th algorithm are significantly different if:
\begin{equation}
\label{eq:posthoc}
 |\bar{R}_i-\bar{R}_j| \geq z^*\sqrt{\frac{m(m+1)}{6n}}.
\end{equation}
where $\bar{R}_i=\tfrac{1}{n}R_{i}$ is the mean rank of the $i$-th algorithm 
and $z^*$ is the Bonferroni corrected $\alpha/m(m-1)$ upper standard normal quantile  \cite[Sec. 12.2.1]{gibbons2011nonparametric}. 
Equation (\ref{eq:posthoc}) is based on the large sample ($n>10$) approximation of the distribution of the statistic.
The actual distribution of the statistic $ |\bar{R}_i-\bar{R}_j| $  is derived assuming
all the $(m!)^n$ ranks in (\ref{eq:friedmanmatrix}) to be equally probable. Under this assumption the variance of $ |\bar{R}_i-\bar{R}_j| $ 
is $m(m+1)/6n$, which originates the term under the square root  in  (\ref{eq:posthoc}).

The sampling distribution of the statistic $|\bar{R}_i-\bar{R}_j|$ assumes all ranks configurations in (\ref{eq:friedmanmatrix}) to be equally probable. 
Yet this assumption is not tenable: the post-hoc analysis is performed \textit{because}
the null hypothesis of the Friedman test has been rejected.
%

\section{Inconsistencies of the mean-ranks test}
We illustrate the inconsistencies  the mean-ranks test 
by presenting three examples.
All examples refer to the analysis of the accuracy of different classifiers on multiple data sets.
We show that the outcome of the test depends both on the actual difference of accuracy between algorithm A and B \textit{and} on the 
accuracy of the remaining algorithms.
\subsection{Example 1: artificially increasing power}
Assume we have tested five algorithms $A,B,C,D,E$ on 20 datasets obtaining the accuracies:
\begin{equation*}
\label{eq:ex2acc}
\begin{array}{c|cccccccccccccccccccc}
\multicolumn{21}{c}{~~Datasets~~}\\
\hline
A & 50 & 50 & 50 & 50 & 50 & 50 & 50 & 50 & 50 & 50 & 80 & 80 & 80 & 80 & 80 & 80 & 80 & 80 & 80 & 80\\
B & 80 & 80 & 80 & 80 & 80 & 80 & 80 & 80 & 80 & 80 & 50 & 50 & 50 & 50 & 50 & 50 & 50 & 50 & 50 & 50  \\
C & 55 & 55 & 55 & 55 & 55 & 55 & 55 & 55 & 55 & 55 & 45 & 45 & 45 & 45 & 45 & 45 & 45 & 45 & 45 & 45 \\
D & 60 & 60 & 60 & 60 & 60 & 60 & 60 & 60 & 60 & 60 & 85 & 85 & 85 & 85 & 85 & 85 & 85 & 85 & 85 & 85 \\
E & 65 & 65 & 65 & 65 & 65 & 65 & 65 & 65 & 65 & 65 & 90 & 90 & 90 & 90 & 90 & 90 & 90 & 90 & 90 & 90\\
\end{array}
\end{equation*} 

The corresponding ranks are:
\begin{equation*}
\label{eq:ex2alg}
\begin{array}{c|cccccccccccccccccccc}
\multicolumn{21}{c}{~~Datasets~~}\\
\hline
A &1  & 1 & 1 & 1 &1 &1  & 1 & 1 & 1 &1 & 3 & 3 &3 &3 & 3 & 3 & 3 &3 &3 & 3\\
B & 5 & 5 &5 &5 & 5 & 5 & 5 &5 &5 & 5 &2  & 2 & 2 & 2 &2 &2  & 2 & 2 & 2 &2\\
C & 2 & 2 &2 &2 & 2 & 2 & 2 &2 &2 & 2 &1  & 1 & 1 & 1 &1 &1  & 1 & 1 & 1 &1 \\
D & 3 & 3 &3 &3 & 3 & 3 & 3 &3 &3 & 3 &4  & 4 & 4 & 4 &4 &4  & 4 & 4 & 4 &4 \\
E & 4 & 4 &4 &4 & 4 & 4 & 4 &4 &4 & 4 &5  & 5 & 5 & 5 &5 &5  & 5 & 5 & 5 &5\\
\end{array}
\end{equation*} 
where better algorithms are given higher ranks. 
We aim at comparing  $A$ and $B$. 
Algorithm $B$ is better than $A$ in the first ten datasets, while $A$ is better than $B$ in the remaining ten.
The two algorithms  have the same mean performance and their differences are symmetrically distributed.
Each algorithms wins on half the data sets.
Different types of two-sided tests (t-test, Wilcoxon signed-rank test, sign-test) return the same $p$-value, $p=1$.
The mean-ranks test correspond in this case to the sign-test and thus also its p-value is 1.
This is most extreme result in favor of the null hypothesis.

Now assume that we compare $A,B$ together with $C,D,E$. 
In the first ten datasets, algorithm $A$ is worse than  $C,D,E$, which in turn are worse than $B$.
In the remaining ten datasets, $C$ is worse than $A,B$, which in turn are worse than $D,E$.
The $p$-value of the Friedman test is $p\approx 10^{-10}$ and, thus, it rejects the null hypothesis.
We can thus perform the post-hoc test (\ref{eq:posthoc}) with $z^*=2.807$
(the Bonferroni corrected $\alpha/m(m-1)$ upper standard normal quantile for $\alpha=0.05$ and $m=5$).
The significance level has been adjusted to $\alpha/m(m-1)$, since we are performing $m(m-1)/2$
two-sided comparisons. The mean ranks of $A,B$ are respectively $2$ and $3.5$ and, thus,
since $|\bar{R}_A-\bar{R}_B|=1.5$ and $z^* \sqrt{\tfrac{m(m+1)}{6n}}\approx 1.4$ we can reject the null hypothesis.
The result of the post-hoc test is that the algorithms $A,B$ have significantly different performance.

The decisions of the mean-ranks test are not consistent: 
\begin{itemize}
 \item if it compares $A,B$ alone, it does not reject the null hypothesis;
 \item if it compares $A,B$ together with $C,D,E$, it rejects the null hypothesis concluding that $A,B$ have significantly different performance.
\end{itemize}
The presence of $C,D,E$ artificially introduces a difference between $A,B$ by changing the mean ranks of $A,B$.
For instance, $D$ and $E$ rank always better than $A$, while they never outperform $B$ when it works well (i.e., datasets from one to ten); 
in a real case study, a similar result would probably indicate that while $B$ is well suited for the first ten datasets, $D,E$ and $A$ are better suited for the last ten. 
The difference (in rank) between $A$ and $B$ is artificially amplified by the presence of $D$ and $E$ only when $B$ is better than $A$. 
The point is that a large differences in the global ranks of two classifiers does not necessarily correspond to large differences in their accuracies (and viceversa, as we will see in the next example). 

This issue can happen in practice.\footnote{We thank the anonymous reviewer for suggesting this example.}
Assume that a researcher presents a new algorithm $A_0$ and some of its weaker variations 
$A_1$, $A_2$,...,$A_k$ and compares the new algorithms with an existing algorithm $B$.
When $B$ is better, the rank is $B\succ A_0 \succ \ldots \succ A_k$. 
When $A_0$ is better, the rank is $A_0 \succ A_1 \succ \ldots \succ A_k \succ B$. 
Therefore, the presence of $A_1$, $A_2$,...,$A_k$ artificially increases the difference between $A_0$ and $B$.


\subsection{Example 2: low power due to the remaining algorithms}

Assume the performance of algorithms $A$ and $B$ on different data sets to be normally distributed as follows:
$$
A\sim N(0,1), ~~B\sim N(1.5,1).
$$

The pool of algorithms comprises also $C,D,E$, whose performance is distributed as follows:
$$
C\sim N(5,1),~~D\sim N(6,1),~~E\sim N(7,1).
$$
A collection of $20$ data sets is considered. 

For the sake of simplicity, assume we want to compare only $A$ and $B$.
There is thus no need of correction for multiple comparisons.

When comparing $A$ and $B$, the power of the two-sided sign test with $\alpha=0.05$ is \textit{very} high: $0.94$
(we have evaluated the power numerically by Monte Carlo simulation).
The power of the mean-ranks test is instead only $0.046$. 
We can explain the large difference of power as follows.
The sign test (under normal approximation of the distribution of the statistic) claims significance when:
$$
|\bar{R}_A-\bar{R}_B|\geq z^*\sqrt{\frac{1}{n}}
$$
while the mean-ranks test (\ref{eq:posthoc}) claims significance when:
$$
|\bar{R}_A-\bar{R}_B|\geq z^*\sqrt{\frac{m(m+1)}{6n}}=z^*\sqrt{\frac{5}{n}},
$$
with $m=5$.
Since the algorithms $C,D,E$ have mean performances that are much larger
than those of $A,B$, the mean-ranks difference $|\bar{R}_A-\bar{R}_B|$ is  equal for the two test.
However the mean-ranks estimates
the variance of the statistic $|\bar{R}_A-\bar{R}_B|$ to be
five times larger compared to the sign test.
The critical value of the mean-ranks test is inflated by $\sqrt{5}$, largely decreasing
the power of the test.
In fact for the mean-ranks test
the variance of $|\bar{R}_A-\bar{R}_B|$ increases with the number of algorithms included in the initial experiment.



\subsection{Example 3: real classifiers on UCI data sets}
Finally, we compare the accuracies of seven classifiers on 54 datasets.
The classifiers are:
J48 decision tree ($C_1$); hidden naive Bayes ($C_2$);  averaged one-dependence estimator (AODE) ($C_3$);  naive-Bayes ($C_4$); J48 graft ($C_5$),
locally weighted naive-Bayes ($C_6$), random forest ($C_7$). 
The whole set of results is given in Appendix.
Each classifier has been assessed via  10 runs of 10-folds cross-validation.
We performed all the experiments using WEKA.\footnote{\url{http://www.cs.waikato.ac.nz/ml/weka/}} 
All these classifiers are described in \citep{witten2005data}.

The accuracies are reported in Table \ref{tab:real}.
Assume that our aim is to compare $C_1,C_2,C_3,C_4$ alone. Therefore, we consider just the first 4 columns in  Table \ref{tab:real}.
The mean ranks are:
$$
C_2=2.676, ~~C_4=1.917, ~~C_1=2.518, ~~C_3=2.888.
$$
The  Friedman test rejects the null hypothesis. The pairwise comparisons for the pair $C_2,C_4$ gives the statistic 
 $$
 z=|\bar{R}_2-\bar{R}_4|/\sqrt{m(m+1)/6n}=3.06.
 $$
  Since $3.06$  is greater than $z^*=2.64$ (the Bonferroni corrected  $\alpha/m(m-1)$ upper standard normal quantile for $\alpha=0.05$ and $m=4$), 
  the mean-ranks procedure finds the algorithms $C_2,C_4$ to be significantly different.

If we compare $C_2,C_4$  together with  $C_1,C_5$, the mean ranks are:
 $$
C_2= 2.713, ~~C_4=2.102, ~~C_1=2.528, ~~C_5=2.657.
$$
 Again, Friedman test rejects the null hypothesis. The pairwise comparisons for the pair $C_2,C_4$ gives the statistic 
 $$
 z=|\bar{R}_2-\bar{R}_4|/\sqrt{m(m+1)/6n}=2.46,
  $$
  which is smaller than $z^*$. Thus  the difference between algorithms $C_2$ and $C_4$ is \textit{not} significant.
  
  The accuracies of $C_2$ and $C_4$ are the same in the two cases but again 
the decisions of the mean-ranks are conditional to the group of classifiers we are considering.

Consider building a set of four classifiers $\{C_2,C_4,C_x,C_y\}$.
By differently choosing $C_x$ and $C_y$ we  can build ten different such sets.
For each subset we run the mean-ranks test to check whether the difference between 
$C_2$ and $C_4$ is significantly different.
The difference is claimed to be \textit{significant}
in $7$ cases and 
\textit{not significant} in $3$ cases.

Now consider a set of five classifiers $\{C_2,C_4,C_x,C_y,C_z\}$. 
By differently choosing $C_x$, $C_y$ and $C_z$  we  can build ten different such sets.
This yields 10 further cases  in which we compare again $C_2$ and $C_4$.
Their difference is claimed to be significant in 9/10 cases. 

Table \ref{tab:contr} reports the pairwise comparisons for which the statistical decision 
changes with the  pool of classifiers that are considered. 
The outcome of the mean-ranks test when comparing the same pair of classifiers 
clearly depends on the pool of alternative classifiers $\{C_x,C_y,\ldots\}$  which is assumed.

\begin{table}
\centering
 \begin{tabular}{cccc}\toprule
 & Card=2 & Card=3 & Card=4\\ \midrule
   \text{$C_2$ vs. $C_4$} & 7/10 & 9/10 & 3/5 \\
   \text{$C_2$ vs. $C_7$} & 1/10 & - & -\\
      \text{$C_3$ vs. $C_7$} & 2/10 & - & -\\
            \text{$C_4$ vs. $C_6$} & 9/10 &5/10 & -\\ \bottomrule
 \end{tabular}
 \caption{Pairwise comparisons that are affected (numbers of decisions that are significantly different/number of subsets) by the performance of the other algorithms.
 Here Card=2 means that, for each pair $C_a,C_b$ on the left column, we are considering the subsets $\{C_a,C_b,C_x,C_y\}$, Card=3 $\{C_a,C_b,C_x,C_y,C_z\}$ and Card=4 $\{C_a,C_b,C_x,C_y,C_z,C_w\}$.  The symbol ``-'' means  that the comparison does not depend on the subset of algorithms.} 
\label{tab:contr}
\end{table} 

\subsection{Maximum type I error}
A further drawback of the  mean-ranks test which has not been discussed in the previous examples is that it  cannot control the maximum type I error, that is, the probability of falsely declaring any pair of algorithms to be different regardless of the other $m-2$ algorithms. If  the accuracies of all algorithms but one are equal, it does not guarantee the family-wise Type I error to be smaller than $\alpha$ when comparing the $m-1$ equivalent algorithms. We point the reader to \citep{Fligner1984}
for a detailed discussion on this aspect.

\section{A suggested procedure}

Given the above issues, we recommend to avoid the mean-ranks test for the post-hoc analysis. One should instead 
perform the multiple comparison
using tests whose decision depend only on the two algorithms being  compared, such as 
the sign test or the Wilcoxon signed-rank test.
The sign test is more robust, as it only assumes the observations to be identically distributed.
Its drawback is low power.
The Wilcoxon signed-rank test is more powerful and thus it is generally recommended \citep{demvsar2006statistical}.
Compared to the sign test, the Wilcoxon signed-rank test makes the additional assumption of a symmetric 
distribution of the differences between the two algorithms being compared. 
The decision between sign test and signed-rank test thus depends on whether the symmetry assumption is tenable 
on to the analyzed data.

Regardless the adopted test,  the multiple comparisons should be performed adjusting the significance level to control the family-wise Type-I error. This can be done using the correction for multiple comparison discussed by
\citep{demvsar2006statistical,garcia2008extension}.
 If we adopt the Wilcoxon signed-rank test in Example 3 for comparing $C_2,C_4$, we obtain the $p$-value $0.0002$, independently from the performance of the other algorithms. 
 Thus, for any pool of algorithms  $C_2,C_4,C_x,C_y$, we always report the same decision:
 $C_2,C_4$ are significantly different because the $p$-value is less 
 than the Bonferroni corrected significance level $\alpha/m(m-1)$ (in the case $m=4$, $\alpha/m(m-1)= 0.0042$).



%

\section{Software}
The MATLAB scripts of the above examples can be downloaded from \url{ipg.idsia.ch/software/meanRanks/matlab.zip}

\section{Conclusions}
The mean-ranks post-hoc test is 
 widely used test for multiple pairwise comparison.
We discuss a number of drawbacks of this test, which
we recommend to avoid.
We instead recommend to adopt the sign-test or the Wilcoxon signed-rank, whose decision does not depend
on the pool of classifiers included in the original experiment.

We moreover bring to the attention of the reader  the Bayesian counterparts of these tests,
which overcome the many drawbacks \cite[Chap.11]{kruschke2010bayesian} of null-hypothesis significance testing.

\bibliography{biblio}
\bibliographystyle{ieeetr}      
\newpage
\appendix
\section*{Table of accuracies used in example 3}
\begin{table}[htp]
\centering
 {
 \fontsize{9}{9}
 \selectfont
\begin{filecontents*}{accuracy.csv}
Dataset,C1,C2,C3,C4,C5,C6,C7
anneal,98.44,98,98,96.43,98.55,98.33,99
audiology,78.32,73.42,71.66,71.23,78.32,77.41,73.89
wisconsin-breast-cancer,93.7,96.71,96.99,97.14,93.7,97.28,95.57
cmc,50.71,52.81,51.39,51.05,50.78,50.98,48.67
contact-lenses,81.67,68.33,71.67,71.67,81.67,65,78.33
credit,86.38,84.64,86.67,86.23,86.52,87.25,85.07
german-credit,72.4,76.6,76.6,76,72.4,75.3,73
pima-diabetes,73.7,74.09,75.01,74.36,73.56,74.75,72.67
ecoli,81.52,80.04,81.83,82.12,81.52,80.63,78.84
eucalyptus,64.28,63.2,58.71,51.1,64.01,59.52,59.4
glass,71.58,74.26,73.83,70.63,71.1,75.69,73.33
grub-damage,38.79,36.88,43.92,47.79,39.42,40.13,42.63
haberman,72.87,71.53,72.52,72.52,72.87,73.52,72.16
hayes-roth,60,56.88,60,60,60,60,59.38
cleeland-14,78.82,81.47,81.8,83.44,78.48,82.78,81.81
hungarian-14,78.64,84.39,84.39,84.74,78.64,84.38,81.97
hepatitis,79.46,85.13,83.79,82.5,79.46,82.5,81.25
hypothyroid,99.28,99.18,98.54,98.3,99.28,98.62,98.97
ionosphere,91.17,90.88,90.88,89.17,91.74,89.17,91.75
iris,93.33,92,92.67,92.67,93.33,92,93.33
kr-s-kp,99.44,92.46,91.24,87.89,99.37,91.21,98.87
labor,85,88,84.67,83,85,81.33,84.67
lier-disorders,56.25,56.25,56.25,56.25,56.25,56.25,56.25
lymphography,78.33,85,85.71,84.38,79,86.33,79.62
monks1,98.74,100,85.44,74.64,98.74,82.21,98.56
monks3,98.92,97.84,96.75,96.39,98.92,96.39,97.84
monks,64.72,64.57,63.73,62.24,64.72,64.9,70.72
mushroom,100,99.96,99.95,95.83,100,99.84,100
nursery,97.05,94.28,92.71,90.32,97.08,91.61,98.09
optdigits,78.97,96.17,96.9,92.3,81.01,94.2,91.8
page-blocks,96.62,96.84,96.95,93.51,96.66,94.15,96.97
pasture-production,75,85.83,80.83,80.83,75,81.67,75.83
pendigits,89.05,97.61,97.82,87.78,89.87,94.81,95.67
postoperatie,70,67.78,67.78,66.67,70,66.67,60
primary-tumor,40.11,48.08,47.49,46.89,40.11,49.55,38.31
segment,94.24,96.36,94.5,91.3,94.03,94.29,96.06
solar-flare-C,88.86,88.24,88.54,86.08,88.86,87.92,86.05
solar-flare-m,90.1,87.02,87.92,87,90.1,86.99,85.46
solar-flare-X,97.84,97.53,97.84,93.17,97.84,94.41,95.99
sonar,74.48,79.83,81.26,80.29,74.45,80.79,78.36
soybean,92.39,94.58,93.4,92.08,92.98,93.55,92.68
spambase,92.81,92.31,93.37,89.85,93.22,90.63,93.65
spect-reordered,78.29,82.07,80.93,79.03,78.29,83.15,80.56
splice,94.36,96.18,96.21,95.36,94.2,95.89,89.37
squash-stored,70,58,60,61.67,70,63.67,57.67
squash-unstored,76.67,69,70.67,61.67,76.67,68.67,77.33
tae,47,44.38,47,47,47,47,45.67
credit,84.93,83.91,85.07,84.2,84.93,85.22,83.33
owel,76.67,84.65,77.78,60.3,76.87,77.88,84.95
waveform,74.38,84.52,84.92,79.86,74.9,83.62,79.68
white-clover,56.9,79.29,68.57,66.9,56.9,64.76,70
wine,88.79,98.33,98.33,98.89,89.35,98.33,97.22
yeast,57.01,57.48,56.74,56.8,57.01,57.48,56.26
zoo,92.18,100,95.09,93.18,92.18,96.18,95.09
\end{filecontents*}
\csvautobooktabular{accuracy.csv}}
\caption{Accuracy of classifiers on different data sets.} \label{tab:real}
\end{table} 




\end{document}